\documentclass{article}

\usepackage{arxiv}

\usepackage[utf8]{inputenc} % allow utf-8 input
\usepackage[T1]{fontenc}    % use 8-bit T1 fonts
\usepackage{hyperref}       % hyperlinks
\usepackage{url}            % simple URL typesetting
\usepackage{booktabs}       % professional-quality tables
\usepackage{amsfonts}       % blackboard math symbols
\usepackage{nicefrac}       % compact symbols for 1/2, etc.
\usepackage{microtype}      % microtypography
\usepackage{lipsum}		% Can be removed after putting your text content
\usepackage{graphicx}
\usepackage{natbib}
\usepackage{doi}

\usepackage{amsmath,amssymb}
\usepackage{xcolor}
\usepackage{xspace}
\usepackage{caption,subcaption} % for subfigures
\usepackage{stmaryrd} % for Iverson brackets
\makeatletter
\DeclareRobustCommand\onedot{\futurelet\@let@token\@onedot}
\def\@onedot{\ifx\@let@token.\else.\null\fi\xspace}

\def\iid{{i.i.d}\onedot}
\def\eg{{e.g}\onedot} 
\def\ie{{i.e}\onedot}

\makeatother

\def\method{\emph{LIMES}\xspace}
\def\single{\emph{incremental}\xspace}
\def\multi{\emph{ensemble}\xspace}
\def\restart{\emph{restart}\xspace}
\def\random{\emph{random}\xspace}
\def\Random{\emph{Random}\xspace}

\def\Single{\emph{Incremental}\xspace}
\def\Multi{\emph{Ensemble}\xspace}
\def\Restart{\emph{Restart}\xspace}

\DeclareMathOperator*{\argmin}{\operatorname{argmin}}
\DeclareMathOperator*{\argmax}{\operatorname{argmax}}

\title{Lightweight Conditional Model Extrapolation\\ for Streaming Data under Class-Prior Shift}

\author{ \href{https://orcid.org/0000-0001-6767-1018}{\includegraphics[scale=0.06]{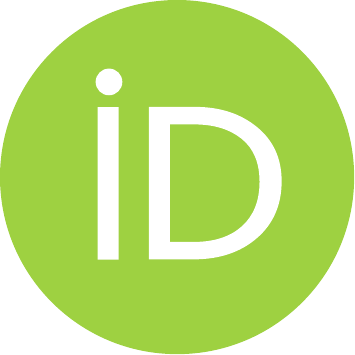}\hspace{1mm}Paulina Tomaszewska}\\
	Faculty of Mathematics and Information Science\\
	Warsaw University of Technology\\
	Warsaw, Poland \\
	\And
	\href{https://orcid.org/0000-0001-8622-7887}{\includegraphics[scale=0.06]{orcid.pdf}\hspace{1mm}Christoph H.~Lampert} \\
	Machine Learning and Computer Vision Group\\
	Institute of Science and Technology Austria (ISTA)\\
    Klosterneuburg, Austria\\
}

\date{}

\renewcommand{\headeright}{}
\renewcommand{\undertitle}{}
\renewcommand{\shorttitle}{\method}

\hypersetup{
pdftitle={Lightweight Conditional Model Extrapolation for Streaming Data under Class-Prior Shift},
pdfsubject={cs.LG, stat.ML},
pdfauthor={Paulina Tomaszewska, Christoph H. Lampert},
pdfkeywords={continual learning, non-stationary data, incremental training, meta-learning, class-prior shift},
}

\begin{document}

\maketitle

\begin{abstract}
We introduce \method, a new method for learning with non-stationary streaming data, inspired by the recent success of meta-learning. The main idea is not to attempt to learn a single classifier that would have to work well across all occurring data distributions, nor many separate classifiers, but to exploit a hybrid strategy: we learn a single set of model parameters from which a specific classifier for any specific data distribution is derived via classifier adaptation. Assuming a multi-class classification setting with class-prior shift, the adaptation step can be performed analytically with only the classifier’s bias terms being affected. 
Another contribution of our work is an extrapolation step that predicts suitable adaptation parameters for future time steps based on the previous data. 
In combination, we obtain a lightweight procedure for learning from streaming data with varying class distribution that adds no trainable parameters and almost no memory or computational overhead compared to training a single model. 
Experiments on a set of exemplary tasks using Twitter data show that \method achieves higher accuracy than alternative approaches, especially with respect to the~relevant real-world metric of lowest within-day accuracy.
\end{abstract}


\section{Introduction}
In our current era of connected devices, data is often generated in a continuous manner, \eg by mobile devices, intelligent sensors or surveillance equipment. 
This creates a~challenge for machine learning systems, especially when the available computational and memory resources are limited, such as when learning \emph{at the edge}. 
On the one hand, systems cannot store all prior data as the necessary amount of storage would grow linearly over the runtime of the system. On~the~other hand, the systems have to make predictions continuously, interleaved with learning, rather than following the traditional \emph{offline} setting with separate training and prediction phases.
Furthermore, continuously generated data is typically non-stationary: its data  distribution varies over time, \eg due to changing environmental conditions or changing user behavior. 

For a machine learning system, a time-varying data distribution is problematic, because it means that previously observed data might not be representative of future data. 
Consequently, simply learning a single fixed model, as is usually done in the standard setting of independent and identically distributed (\iid) data, leads to suboptimal results.
Alternatively, it is possible to learn multiple models, essentially starting a new one whenever the data distribution changes. 
This, however, can also lead to suboptimal results when each model achieves acceptable prediction quality only after having seen  enough data. Furthermore, learning multiple models leads to increased memory requirements.

In this work, we introduce \method (short for \underline{Li}ghtweight \underline{M}odel 
\underline{E}xtrapolation for \underline{S}treaming data)
that offers a new way for learning classifiers from 
non-stationary streaming data in resource-constrained 
settings. 
Inspired by recent progress in \emph{meta-learning}, 
\method's core idea is to combine the resource efficiency of 
having only a single set of parameters with the flexibility 
of learning different models for different data distributions.
We construct a base model that can be adapted efficiently to different target distributions together with a procedure for forecasting the parameters necessary for adapting the base model to be used on future data, even before any examples of that data have been observed.

Specifically, we adopt the scenario of multi-class classification 
under \emph{class-prior shift}, in which the class frequencies
change over time, while the class-conditional data densities 
remain approximately constant. 
One of our contributions is to show how in this setting it is possible to learn a model 
from multiple data distributions, and how make use of it for yet another data distribution, with hardly any computational or memory overhead.
Furthermore, we introduce an \emph{extrapolation} step that forecasts the class distribution of future data based on the observed previous data. % recent history. 

We show experimentally on exemplary tasks using Twitter data that 
\method achieves higher accuracy than either learning a single model or an ensemble of multiple models. The code is available at \texttt{https://github.com/ptomaszewska/LIMES}.

\section{Method}
In this section, we formalize the learning scenario, we study and introduce the proposed \method method.

\subsection{Learning Scenario}
We assume a time-varying data distribution $p_t(x,y)$ for $t=1,2,\dots$, where $x\in \mathcal{X}$ 
are the designated inputs to the~learning system and $y\in \mathcal{Y} =\{1,\dots,L\}$ are the output labels~(\cite{bartlett1992learning}). 
The goal of a learning system is to create a~predictive model, 
$f_t:\mathcal{X}\to\mathcal{Y}$, for any time step $t$, with as 
small as possible \emph{risk}, $R_t(f)=\mathbb{E}_{(x,y)\sim p_t}[\ell(y,f_t(x))]$, 
where $\ell$ is the $0/1$-loss function, $\ell(y,y')=\llbracket y\neq y'\rrbracket$\footnote{The \emph{Iverson} brackets $\llbracket P\rrbracket$ for a predicate $P$ take the value $1$ if 
$P$ is true and $0$ otherwise.}
For this task, the learning system can make use of observed 
data from prior time steps, $S^1$, \dots, $S^{t-1}$, where 
each  $S^{\tau}=\{(x^\tau_1,y^\tau_1),\dots,(x^\tau_{n_\tau},y^\tau_{n_\tau})\}\sim p_\tau(x,y)$, for $\tau=1,\dots,t-1$.
In line with practical resource-constrained applications, we assume a continuously running learning system that is kept up-to-date by \emph{periodic retraining}~(\cite{graepel2010web,mcmahan2013ad,li2015click}).
That means, only at the end of each time step, it is  possible to update the model parameters. 
The obtained model is then used to make predictions for all data during the next time step.

A typical example of this setting is the real-time classification of social media streams.
For example, for our experimental evaluation in Section~\ref{sec:experiments}, the inputs are tweets and the~outputs are the countries from which the tweets originated. 
The~time index corresponds to the time intervals over which the data distribution can be assumed to be relatively stable, in our case \emph{hours}. 
Within a few time steps, the data distribution can change quite rapidly.
Intuitively, users from different countries tweet predominantly at different times due to their respective time zones. 
This causes a time-varying data distribution in which class proportions fluctuate substantially over time. 

\subsection{Learning Methods}\label{sec:methods}
A number of possible approaches can be used to tackle the learning problem introduced above.
The most classical is standard \emph{empirical risk minimization}~(\cite{vapnik2013nature}),
in which a model, $f$, from a class
of possible ones, $\mathcal{F}$, is sought that minimizes the training error across all the data observed so far, 
\begin{align}
f_{t+1} &= \argmin_{f\in\mathcal{F}} \sum_{\tau=1}^{t} \hat{R}_{\tau}(f),\label{eq:erm}
\intertext{where}
\hat{R}_{\tau}(f)&=
\sum_{(x,y)\in S^{\tau}}\!\!\ell(y,f(x))
\end{align}
is the training error (empirical risk) of a model $f$ on the data from the time step $\tau$. 
Optionally, a regularizer or other terms can be added to improve generalization or to make the optimization problem better behaved. 

Empirical risk minimization is widely used when data is sampled \iid from a stationary distribution. However, for learning from non-stationary data, it has a number of shortcomings.
First, Equation~\eqref{eq:erm} will not actually learn a model adapted for the coming time step $t+1$, but rather one that works well on average across all data distributions in the past. 
Second, minimizing Equation~\eqref{eq:erm} is intractable in a practical streaming setting.
It would require having stored all training sets, $S^1,\dots,S^{t}$. which is impractical 
for systems that are meant to run continuously over a long time with limited resources, 
such as \emph{at the edge}~(\cite{murshed2021machine}). 

Alternatively, one can find a model by training only on the most recently available data, 
\begin{align}
f_{t+1} &= \argmin_{f\in\mathcal{F}} \hat{R}_{t}(f).\label{eq:recent}
\end{align}
This \emph{restart} approach overcomes the second problem and mitigates the first one. 
However, the procedure completely ignores non-recent data. Therefore, the models it produces are usually limited in the accuracy they can achieve.

\emph{Incremental training}~(\cite{giraud2000note}) avoids the problem of having 
to store all data, but still allows models to benefit from all available data.
Each $f_{t+1}$ is obtained by running a minimization routine on the most recent empirical risk, $\hat{R}_{t}$, as in Equation~\eqref{eq:recent}.
However, instead of running the optimization routine until 
convergence, one initializes 
the process at the parameters  
of the previous model, 
$f_{t}$, and performs early 
stopping, \eg after one pass 
through the data. 
Models learned by incremental 
training depend on all available 
data. At the same time, they 
tend to be biased towards recently 
seen data. The latter can be a 
benefit if the data distribution 
changes slowly, but it can also 
be a disadvantage, if data from 
more distant time steps is 
valuable, such as for periodic data 
distributions.

A way to preserve information from more distant time step
is to combine incremental training with \emph{model ensembling}~(\cite{krawczyk2017ensemble,street2001streaming,minku2011ddd,muhlbaier2007multiple}).
Instead of training a single classifier, one trains multiple prototypical models, $g_1,\dots,g_K$, on different subsets of the available data. 
The desired $f_{t+1}$ is then constructed out of this ensemble based on contextual information or, if available, data from the target distribution~(\cite{dsc_survey,dsc_drift}). 
The~cost of the gained flexibility is a $K$-fold increase in memory requirements in order to store all models.
At the same time, the amount of training data necessary to achieve satisfactory accuracy tends to be larger as well, when each model is trained only on a subset of all observed data. 

Recently, the problems of \emph{class-incremental learning}~(\cite{rebuffi2017icarl}) and \emph{task-incremental learning}~(\cite{li2017learning}) have attracted a lot of attention in the machine learning community, see \eg~(\cite{delange2021continual}) for a survey.
These problems are special cases of the problem of learning with non-stationary data, based on additional assumptions about which data appears when and to which tasks the resulting functions will be applied.
Specialized algorithms have been derived that aim at giving deep neural networks the~ability to learn continually without suffering from catastrophic forgetting~(\cite{mccloskey1989catastrophic,french1999catastrophic}). 
However, those recent methods are not applicable in the situation in which we are interested. On the one hand, that is because they solve different problems, and on the other hand, because they require additional resources either for storing part of the observed data~(\cite{rebuffi2017icarl,he2020incremental}) or growing the model architecture over time~(\cite{rusu2016progressive,yoon2018lifelong}).  

\subsection{Classifier Adaptation}
The \method method we propose, combines the best aspects of the different approaches discussed before.
Like in incremental training, only a single model 
is trained, for which all available data is used.
Like in ensemble learning, the predictions are made with a model specifically adapted to the target 
distribution, thereby allowing for higher prediction accuracy. 
The core concept to achieve both properties simultaneously is a form of \emph{classifier adaptation}~(\cite{saerens2002adjusting,royer2015classifier}), that we discuss in the following.

We adopt a setting of probabilistic multi-class 
classification, such as implemented by logistic 
regression or any neural networks with softmax 
output layer.
Such a model represents conditional distributions 
over the label set, $\mathcal{Y}=\{1,\dots,L\}$, in log-linear form
\begin{align}
\hat p(y|x;W,b) &\propto \exp({w_y^\top\phi(x)+b_y}). \label{eq:loglinearmodel}
\end{align}
The model parameters are the weight vectors $W=(w_1,\dots,w_L)\in\mathbb{R}^{d\times L}$, and per-class bias terms $b=(b_1,\dots,b_L)\in\mathbb{R}^{L}$. 
The feature map $\phi:\mathcal{X}\to\mathbb{R}^d$ can be fixed (in the case of logistic regression) or learned together with the~classifier parameters (in the case of a neural network). 

The proportionality constant for the right hand side of Equation~\eqref{eq:loglinearmodel} can be computed efficiently for any $x\in\mathcal{X}$ as $\frac{1}{Z(x)}$ with $Z(x)=\sum_{z\in\mathcal{Y}}\exp({w_z^\top\phi(x)+b_z})$. We generally suppress this step from our notation for the sake of conciseness.

The probabilistic model allows one to make predictions for new inputs $x\in\mathcal{X}$ by
\begin{align}
f(x) &= \argmax_{y\in\mathcal{Y}}\hat p(y|x;W,b)
=  \argmax_{y\in\mathcal{Y}}\ [w_y^\top\phi(x)+b_y].
\label{eq:argmaxclassifier}
\end{align}

The parameters $W,b$ (and potentially $\phi$) are typically trained to make the conditional model distribution, $\hat p(y|x;W,b)$, as close as possible to the conditional distribution of the training data, $p(y|x)$.
In the case that they are identical, Equation~\eqref{eq:argmaxclassifier} yields the \emph{Bayes-optimal} prediction rule, \ie the classifier has minimal expected error on future data from the same distribution. 
However, if the data distribution at the prediction time, $q(x,y)$, differs from $p(x,y)$, then the~classifier $f$ can become suboptimal, \ie have higher error rate than necessary.
Unfortunately, the latter situation is the rule rather than the exception in the setting we study. For a time-varying data distribution, the data available at training time practically never is distributed identically to the data at prediction time. 

There is no general solution for this problem of \emph{distribution mismatch}, unless we assume that $p(x,y)$ and $q(x,y)$ are in some way related to each other. 
One such relatedness assumption, which is often justified in practice, is that the~two distributions differ only by a \emph{class-prior shift}, that is, $q(y)\neq p(y)$ but  $q(x|y)=p(x|y)$~(\cite{quinonero2008dataset}).
In that case, it is possible to \emph{adapt} $f$ to a classifier that yields optimal predictions for the new situation.  
To see this, we first write
\begin{align}
    q(y|x) &= \frac{q(x|y)q(y)}{q(x)}\label{eq:adapt}
    = \frac{p(x|y)q(y)}{q(x)} %\label{eq:adapt2}
    = p(y|x)\frac{q(y)}{p(y)}\frac{p(x)}{q(x)}, %\label{eq:adapt3}
\end{align}
where the first and third equality are Bayes' rule and the second equality follows from the class-prior shift assumption.
Consequently, the optimal decisions for 
the distribution $q(x,y)$ is
\begin{align}
    \argmax_{y\in\mathcal{Y}} q(y|x) 
    &= \argmax_{y\in\mathcal{Y}} p(y|x)\frac{q(y)}{p(y)}, \label{eq:adapt4}
\end{align}
where we have used the fact that the $\argmax$ is not affected by multiplication with factors that do not depend on $y$.

For a log-linear model parameterized as in Equation~\eqref{eq:loglinearmodel} %

that approximates $p(y|x)$, we obtain the \emph{adapted model}
\begin{align}
\hat q(y|x;W',b')&\propto \exp({{w'_y}{}^\top\phi(x)+b'_y})\label{eq:adaptedmodel}
\intertext{that approximates $q(y|x)$ by setting, for all $y\in\mathcal{Y}$,}
w'_y=w_y,&\quad b'_y=b_y+\log\frac{q(y)}{p(y)},
\label{eq:adaptedparams}
\end{align}
as long as $q(y)$ and $p(y)$ are known, or can at least be estimated sufficiently well. 
We denote the classifier associated with the adapted model 
\eqref{eq:adaptedmodel} as $f_{p\to q}$.

\subsection{Lightweight Model Extrapolation for Streaming Data}
The ability to adapt a learned model from one data distribution to another has been used previously, for example for \emph{domain adaptation}~(\cite{quinonero2008dataset}) or to exploit changing class proportions at prediction time~(\cite{royer2015classifier}).

In contrast to prior work, we use it at training time to learn a single set of parameters that will work well --after respective adaptation-- under many different conditions.
Conceptually, the~step resembles meta-learning~(\cite{maml,finn2019online}), where also a base model is learned in a way such that it can 
be adapted easily to new situations. 
Specifically, we form a single log-linear model $f_u$ 
with modeled conditional distribution as in Equation~\eqref{eq:loglinearmodel}.

Instead of training this model to minimize 
the average loss across all time steps, we 
aim for the adapted models to result in minimal 
loss, 
\begin{align}
\min_{W,b} &\quad
\sum_{\tau=1}^{t} \hat{R}_{\tau}(f_{u\to p_{\tau}}), \label{eq:meta}
\end{align}
where $u$ denotes the distribution in which all classes have equal probability. 
Despite the similar formulation, learning a model using Equation~\eqref{eq:meta} differs fundamentally from learning a model with Equation~\eqref{eq:erm}. 
For the latter, the terms, $\hat R_{1}(f),\dots,\hat R_t(f)$, in the summation compete with each other: the parameters that optimally minimize one of them are not optimal for any of the others, when their respective training data is distributed differently. As a consequence, the optimization has to settle for finding parameters that perform well on average, but are never truly optimal. 
This effect is purely due to the time-varying nature of the data, so neither observing larger datasets nor more of them will overcome it. 

In the newly proposed formulation~\eqref{eq:meta} the terms, $\hat R_{1}(f_{u\to p_1}),\dots,\hat R_t(f_{u\to p_t})$, support each other:
up to finite sample effects, the parameters that minimize any of the terms $\hat R_{\tau}(f_{u\to p_{\tau}})$ for $\tau=1,\dots,t$, are identical,
and they are also the same ones that would minimize the risk of the~modeled $f_u$ over a dataset sampled from $u$, \ie with uniform class distribution. 
The reason is that the differences in class  distributions are compensated by the corresponding adaptation step, so the~model parameters do not have to change. 
As~a~consequence, the more datasets we observe or the larger they are, the higher the precision with which we can recover the optimal parameters for all time steps.

As in the case of empirical risk minimization, solving Equation~\eqref{eq:meta} exactly is intractable because it would require us to store all past data. 
Instead, we again use incremental training: we 
update the parameters by always minimizing the~most recent empirical risk
\begin{align}
\min_{W,b} &\quad
\hat{R}_{t}(f_{u\to p_{t}}), \label{eq:metacont}
\end{align}
starting at the previous parameter values and using early stopping after a single pass through the data. 

Forming $f_{u\to p_{\tau}}$ for any $\tau=1,\dots,t$, requires knowledge of the class probabilities $p_{\tau}(y)$. 
These are not known exactly, but after having observed the training set $S^{\tau}$, we can estimate them easily by the observed class frequencies, $\hat p_{\tau}(y) = \frac{1}{n_{\tau}}\sum_{i=1}^{n_{\tau}} \llbracket y^{\tau}_i=y\rrbracket$.
Those are just $L$-dimensional vectors, so they can be stored, even continuously, with only a~very small memory footprint.

The model $f_u$ resulting from the above procedure can be expected to work well for data with a uniform class distribution,
and also for any other class distribution if we adapt it suitably.

This property is particularly useful for constructing a model for the next time step, $t+1$: instead of having to train on a data set $S^{t+1}\sim  p_{t+1}$, which is not available yet, we can simply perform the very efficient adaptation step~\eqref{eq:adaptedmodel} to form $f_{u\to p_{t+1}}$. 
This, however, requires an estimate of the~class probabilities at the following time step, $p_{t+1}(y)$, which are a priori unknown. 
To overcome this problem, we propose a history-based \emph{forecasting step} that extrapolates the class distribution.
First, we find the time step, $t^*$, among all earlier ones whose class distribution is 
most similar to the one at the current 
time step, $t$. 

\begin{align}
t^* = \argmin_{\tau=1,\dots,{t-1}}
\text{dist}(\hat p_{\tau}(y),\hat p_{t}(y)),
\end{align}
where $\text{dist}$ denotes the $L^1$-distance.
We then use $\hat p_{t^*+1}$ as a~proxy for $\hat p_{t+1}$.
In other words, we forecast that the situation one time step after the current, $t+1$, will be similar to the situation that had emerged one time step after a similar situation had occurred in the past, $t^*+1$.

The combination of both steps, adaptation at training time and forecasting of the next class distribution, we call \method (Lightweight Model Extrapolation for Streaming Data). 
It has a number of desirable properties. 
First, it is easy to implement, as it requires neither a new model architecture nor a special form of optimization. Instead, one simply trains a model continuously on the incoming data, and only adjusts the model's bias terms at each new time step.
\method's computational and memory overheads are negligible, as only the vectors of prior class distributions have to be stored and searched. 
For very long-running systems, the forecasting step could be made even more efficient (time and memory complexity $O(1)$ instead of $O(t)$) by either using a rolling history buffer, or by selecting a finite number of prototypical histories in a data-dependent way. 
Such steps are orthogonal to our main contribution in this work, so we leave them to future work.

\begin{figure}\centering
    \includegraphics[width=.55\columnwidth]{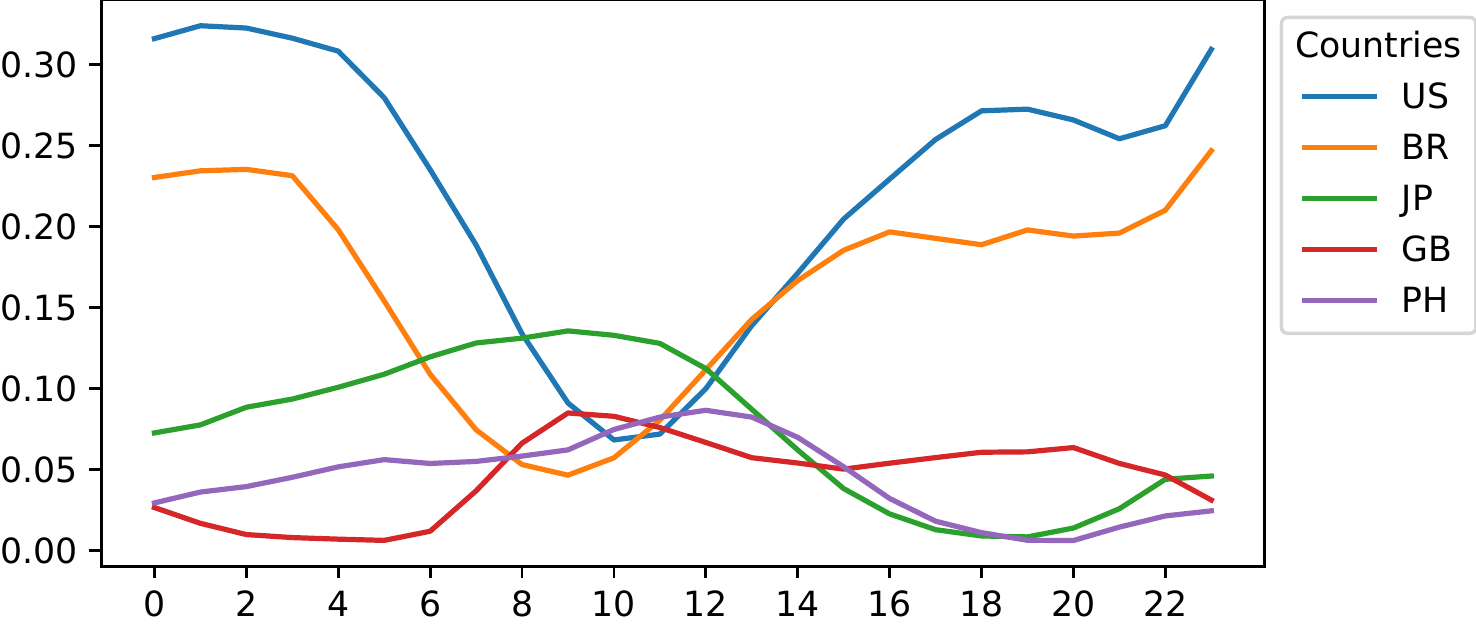}
    \caption{Class distribution in \texttt{geo-tweets} dataset across 24 hours (March 1, 2020) for five most frequent countries (\texttt{US}: USA, \texttt{BR}: Brazil, \texttt{JP}: Japan, \texttt{GB}: Great Britain, \texttt{PH}: Philippines). }
    \label{fig:countries} 
\end{figure}
\section{Experimental Evaluation}\label{sec:experiments}
We evaluate \method on the \texttt{geo-tweets} dataset\footnote{\url{https://cvml.ist.ac.at/geo-tweets}}.
It consists of a large number of tweets with geographic 
information that were collected using Twitter's free 
Streaming API\footnote{\url{https://developer.twitter.com/en/docs/tutorials/consuming-streaming-data}} from April to July 2020.
The associated classification task is to predict from which of 250 countries a tweet was sent based on its textual content and/or the user metadata.
Because users in different time zones are active at different times of the~day, the~country distribution
fluctuates strongly across time.
Figure~\ref{fig:countries} illustrates this effect: 
the $x$-axis shows the hours of a~day in GMT time zone. Between hours 14 and 23 as well as 0 and 6, the most active countries on Twitter are the USA and Brazil. At their peak (hours 22-4), they together are responsible for more than half of all tweets in the dataset.
In~contrast, between hours 7-14, tweets from the USA and Brazil are less frequent, whereas Japan, Great Britain and the Philippines are responsible for larger shares. 

To reflect the time-varying nature of the data, 
we split the~data into hourly chunks, which on 
average contain approximately 127000 tweets 
(minimum: 61000, maximum: 149000). We use 
random subsets of 80\% of each chunk
for training and 20\% for evaluation.
In order to assess statistical variations, we subdivide the data further. First, we create 
two subsets, one containing tweets of the 
1st to 5th day of each month (\emph{"early"})
and one for the 11th to 15th (\emph{"late"}).
From each subset, we create ten realizations
of the time series by subsampling the~data 
with a stride of 10 from different starting 
indices.

From each tweet we extract two 512-dimensional 
feature vectors using the pretrained  \texttt{distiluse-base-multilingual-cased-v1} 
multi-lingual sentence embedding network~(\cite{reimers-2019-sentence-bert}).
One is of the tweet text itself and one is of the text in the \emph{location} meta-data, which is a user-provided free text that might or might not be related to the actual location of the user. 
As target labels, we use the tweet's country of origin, which for geo-tagged tweets can be extracted from the tweet's meta-data.
Specifically, we infer the country either from the lat-long coordinates, or, if those are not provided, we use the user-specified \emph{country\_code} entry of the meta-data. 

The input features can be extracted also for tweets without geo-tags. Therefore, the classifier resulting from the~described training method can be used to predict the~country of origin from any tweet, which is an often-required step in social media analysis~(\cite{han2014text}).
Note, however, that in this work, we are not trying to compete with existing methods for this specific task, as the Twitter data serves just as a real-world testbed for our proposed general-purpose model adaptation technique.

As an instantiation of \method, we implement a logistic regression model in the \emph{keras} toolkit with \emph{tensorflow} backend. 
The training is performed incrementally by Equation~\eqref{eq:meta} using the Adam optimizer with mini-batches of size 100. 
No regularization is used and all hyperparameters 
are left at their defaults 
(learning rate $0.001$, $\beta_1=0.9$, $\beta_2=0.999$).

As baseline, we use alternative methods
discussed in Section~\ref{sec:methods},
relying on the same model architecture
and optimization routines as for \method. 
\Single is a classifier that is trained 
incrementally to minimize~\eqref{eq:erm}, \ie 
identically to \method but without adaptation
and forecasting.
\Random uses the same adaptation as \method, but using a randomly selected time step from history instead of the extrapolation.
\Multi is an ensemble of 24 classifiers, one per hour. 
Each classifier is trained incrementally on the data of the respective hour. For prediction at time step $t+1$, the model with index $(t+1)\mod 24$ is used. 
\Restart trains a new model at every time step 
$t$ from the data $S^{t}$ and uses it for 
predictions at step $t+1$. 

All four methods are tested in 60 experimental
settings: 2 subsets (\emph{early}, \emph{late}), 
3 feature types (\emph{tweet}, \emph{location} 
or their concatenation), and 10 different 
data realization. 
In each case, we measure the accuracy of the 
resulting classifier. We then report average 
and standard deviation across the ten realizations, 
once for the average accuracy across all hours 
(\emph{avg-of-avg}) and once for the average of the 
minimum accuracy within each day (\emph{avg-of-min}).
The latter value is often a quantity of interest 
for real-world systems, because customer satisfaction 
tends to depend not on the average of their experience but rather on the worst case.

\subsection{Results}

Table~\ref{tab:results} summarizes the results. 
\begin{table*}[t]\centering\small
\caption{Experimental Results: classifier accuracy of the proposed \method method and baselines as mean and standard deviation across 10 data splits for six experimental settings (see main text for details). Reported are the average per-day accuracy (\emph{avg-of-avg}) and the minimum-across-the-day accuracy (\emph{avg-of-min}).
The difference of LIMES to the other methods 
is significant according to a Wilcoxon signed
rank test at a below $0.5\%$ level in all cases.
}\label{tab:results} 
\setlength\tabcolsep{5pt}
\begin{subfigure}{.3\textwidth}\centering
\caption{features: \emph{tweet}, subset \emph{0--5} }
\begin{tabular}{c|cc}
\textbf{method}& \textbf{avg}& \textbf{min}\\\hline
LIMES& $52.35_{\pm 0.04}$ & $43.15_{\pm 0.21}$ \\
incremental& $51.81_{\pm 0.05}$ & $40.51_{\pm 0.25}$ \\
random& $48.85_{\pm 0.13}$ & $35.13_{\pm 0.71}$ \\
ensemble& $38.94_{\pm 0.03}$ & $29.48_{\pm 0.18}$ \\
restart& $34.28_{\pm 0.05}$ & $24.62_{\pm 0.27}$ \\
\end{tabular}
\end{subfigure}\quad
\begin{subfigure}{.3\textwidth}\centering
\caption{features: \emph{location}, subset \emph{0--5} }
\begin{tabular}{c|cc}
\textbf{method}& \textbf{avg}& \textbf{min}\\\hline
LIMES& $81.23_{\pm 0.03}$ & $75.20_{\pm 0.07}$ \\
incremental& $81.09_{\pm 0.03}$ & $74.49_{\pm 0.14}$ \\
random& $79.72_{\pm 0.06}$ & $71.84_{\pm 0.50}$ \\
ensemble& $68.48_{\pm 0.05}$ & $62.09_{\pm 0.13}$ \\
restart& $54.33_{\pm 0.05}$ & $44.28_{\pm 0.17}$ \\
\end{tabular}
\end{subfigure}\quad
\begin{subfigure}{.3\textwidth}\centering
\caption{features: \emph{tweet--location}, subset \emph{0--5} }
\begin{tabular}{c|cc}
\textbf{method}& \textbf{avg}& \textbf{min}\\\hline
LIMES& $85.54_{\pm 0.03}$ & $79.92_{\pm 0.11}$ \\
incremental& $85.48_{\pm 0.03}$ & $79.40_{\pm 0.14}$ \\
random& $84.13_{\pm 0.04}$ & $76.80_{\pm 0.19}$ \\
ensemble& $72.37_{\pm 0.04}$ & $66.16_{\pm 0.15}$ \\
restart& $58.36_{\pm 0.04}$ & $48.08_{\pm 0.18}$ \\
\end{tabular}
\end{subfigure}
\vskip.5\baselineskip
\begin{subfigure}{.3\textwidth}\centering
\caption{features: \emph{tweet}, subset \emph{10--15} }
\begin{tabular}{c|cc}
\textbf{method}& \textbf{avg}& \textbf{min}\\\hline
LIMES& $52.15_{\pm 0.04}$ & $42.90_{\pm 0.22}$ \\
incremental& $51.51_{\pm 0.03}$ & $39.90_{\pm 0.22}$ \\
random& $48.63_{\pm 0.13}$ & $34.79_{\pm 0.62}$ \\
ensemble& $38.74_{\pm 0.05}$ & $29.39_{\pm 0.16}$ \\
restart& $33.81_{\pm 0.05}$ & $24.21_{\pm 0.19}$ \\
\end{tabular}
\end{subfigure}\quad
\begin{subfigure}{.3\textwidth}\centering
\caption{features: \emph{location}, subset \emph{10--15} }
\begin{tabular}{c|cc}
\textbf{method}& \textbf{avg}& \textbf{min}\\\hline
LIMES& $81.17_{\pm 0.04}$ & $75.07_{\pm 0.16}$ \\
incremental& $80.98_{\pm 0.04}$ & $74.24_{\pm 0.17}$ \\
random& $79.62_{\pm 0.05}$ & $71.79_{\pm 0.30}$ \\
ensemble& $68.18_{\pm 0.04}$ & $61.30_{\pm 0.22}$ \\
restart& $53.94_{\pm 0.09}$ & $43.34_{\pm 0.39}$ \\
\end{tabular}
\end{subfigure}\quad
\begin{subfigure}{.31\textwidth}\centering
\caption{features: \emph{tweet--location}, subset \emph{10--15} }
\begin{tabular}{c|cc}
\textbf{method}& \textbf{avg}& \textbf{min}\\\hline
LIMES& $85.48_{\pm 0.03}$ & $79.74_{\pm 0.11}$ \\
incremental& $85.36_{\pm 0.02}$ & $79.15_{\pm 0.10}$ \\
random& $84.09_{\pm 0.07}$ & $76.64_{\pm 0.19}$ \\
ensemble& $72.09_{\pm 0.05}$ & $65.60_{\pm 0.25}$ \\
restart& $57.73_{\pm 0.06}$ & $46.53_{\pm 0.26}$ \\
\end{tabular}
\end{subfigure}
\end{table*}
Overall, one can see that the task of predicting 
the country from the tweet text is more difficult 
than from the user-provided meta-data. Combining 
both feature types yields the highest accuracy.
In all cases, the~minimal accuracy across the day 
is substantially lower than the average. This 
indicates that the difficulty of the~classification 
task varies over time, presumably due to changes 
in the label distribution. 

Comparing the different methods, we observe a
ranking that is stable across all experimental 
settings:
\method achieves the best results, followed by
\single and \random. \Multi performs substantially worse, and \restart even worse than that. 
Given that \single differs from \method only in 
the lack of an adaptation and forecasting step, 
we can conclude that the way we propose the adaptation
to the class distribution indeed has a positive effect.
For the \emph{avg-of-avg accuracy} measure, the~gain 
due to adaptation and forecasting is the biggest 
for the most difficult task: $0.53$--$0.65$\% when 
predicting the country based on the tweet text 
itself. 
It gets smaller, the easier the~task becomes:
$0.13$--$0.19$\% for \emph{location} features, and 
$0.07$--$0.12$\% for the concatenation of both.
More apparent, however, is the positive effect 
on the \emph{avg-of-min accuracy}: for \emph{tweet} 
features, it is $2.64$--$3.00$\%, for \emph{location} 
$0.71$--$0.83$\% and for their concatenation 
$0.53$--$0.60$\%. 

\begin{figure*}[t]\centering
\begin{subfigure}{\textwidth}\centering
\includegraphics[height=.24\textwidth]{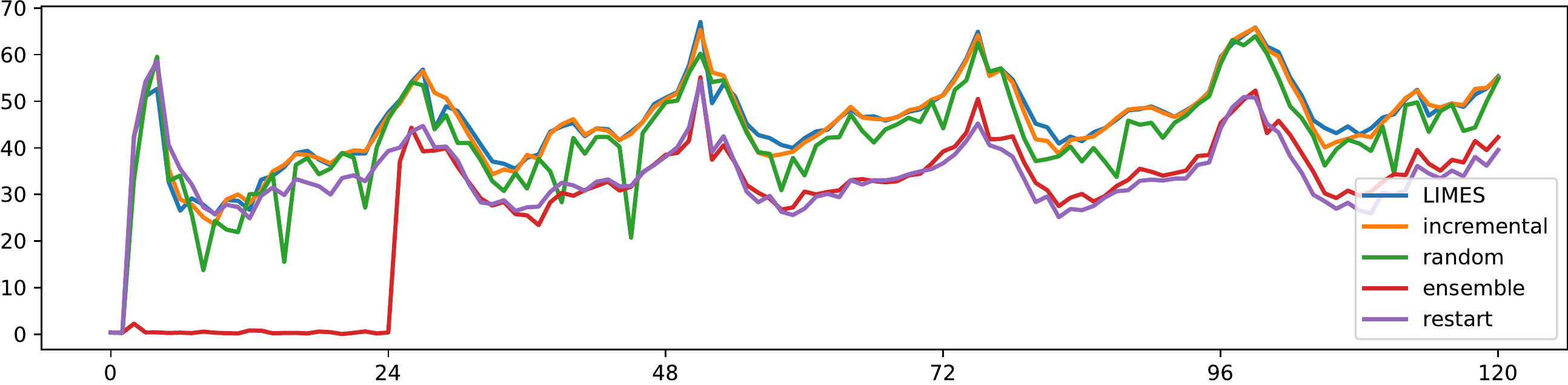}
\caption{Exemplary curve (realization 0) of per-hour accuracy for first 120 hours (\emph{March 1st--5th, 2020).}}
\end{subfigure}
\vskip2\baselineskip
\begin{subfigure}{.46\textwidth}\centering
\includegraphics[height=.52\textwidth]{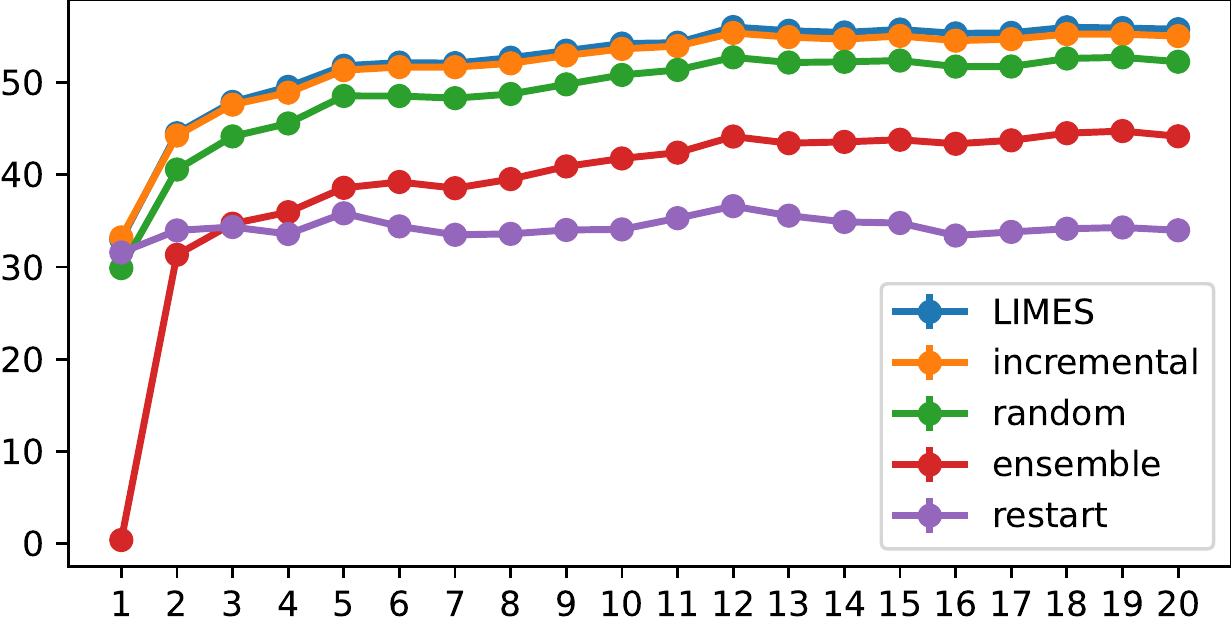}
\caption{Curve of average per-day accuracy over 20 days}
\end{subfigure}
\quad
\begin{subfigure}{.46\textwidth}\centering
\includegraphics[height=.52\textwidth]{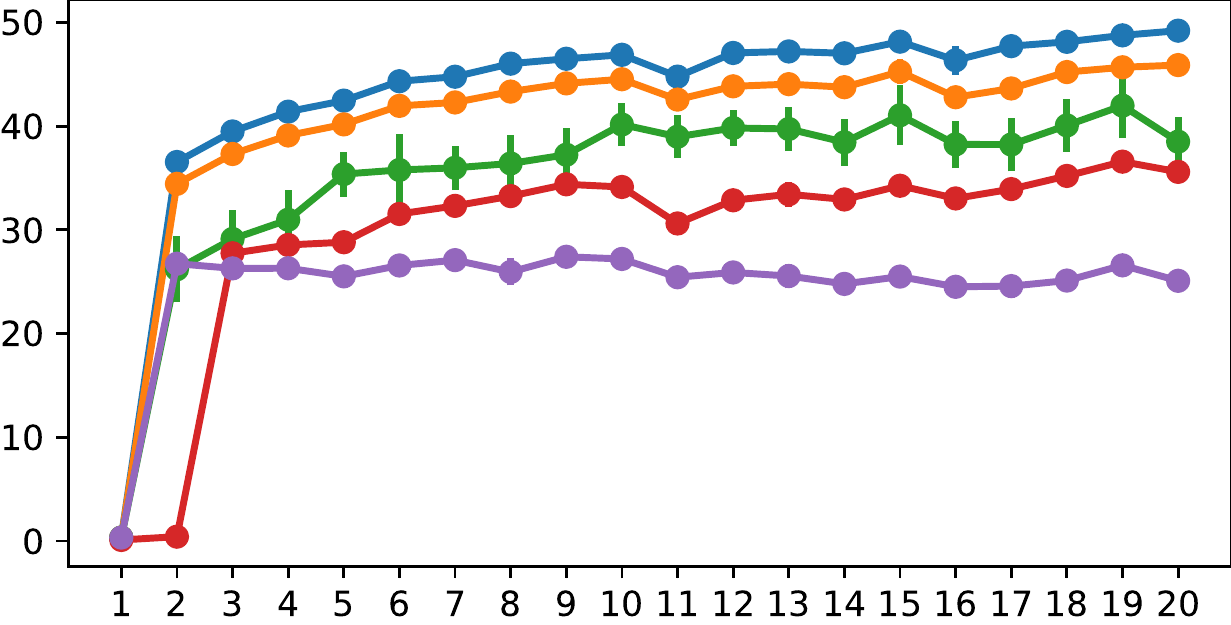}
\caption{Curve of minimum-across-day accuracy over 20 days}
\end{subfigure}
\caption{Classifier accuracy over time for the proposed \method and baseline methods in one of the tested settings (\emph{tweet} 
features, subset \emph{early}). 
}\label{fig:accuracy}
\end{figure*}

Figure~\ref{fig:accuracy} provides an illustrative
explanation of these results. Subfigure (a) shows 
the four methods' accuracy curves over the first 120
hours of one exemplary learning run (\emph{tweet} 
features, subset \emph{early}, realization \emph{0}). 
First, one can clearly see the 24-hour cycle by 
which the accuracies of all classifiers fluctuate. 
The accuracies are highest at around 2:00 GMT. As we
know from Figure~\ref{fig:countries}, this corresponds 
to the time the USA and Brazil are by far the most 
common countries from which tweets originate. 
The accuracy is the lowest around 9:00 GMT, which is the 
time when tweets come from several different countries 
in comparable amounts. 

Second, one can also see that \multi, which does not 
share information between different hours, starts working 
only after 24 hours and afterwards increases 
its prediction quality quite slowly. In contrast, the other models make reasonable predictions already after the first hour,
because they transfer information between different hours.

Finally, one observes that \method is superior over \single mostly in the \emph{valleys} around 9:00 GMT, \ie during the~hours where the classification problem is the hardest. 
Presumably, this is because \single learns parameters that are adapted to the average data distribution, and that is closer to the simple situation with two dominant countries than to the difficult situations where many countries are similarly likely.

Subfigures (b) and (c) make the difference between the~methods more visible by suppressing the daily fluctuations. 
They show the accuracy values for all 20 days aggregated 
as the per-day-average and per-day-minimum,  respectively. The line plots show mean of results over all ten realizations, whereas the error bars reflect the standard deviation. As expected, standard deviation is the biggest in the case of \random method as for each time step, it draws a class distribution vector from historical values. This effect of randomness is alleviated when the average per-day is computed.

At subfigures (b) and (c), one can see the difference between \method and \single well: 
it is small but consistent for the per-day-average accuracy, 
and bigger for the per-day-minimum accuracy.\\

Incremental training is clearly effective, as can be seen 
from the fact that the accuracy curves for the three 
incrementally trained methods, \method, \single and 
\multi, increase over time.
In contrast, \restart's accuracy is approximately constant 
(up to the hourly fluctuations) over the runtime of the 
system. 

\section{Summary and Conclusions}
In this work, we introduced \method, a lightweight method
for continuous classifier training from streaming data with class-prior shift. 
Inspired by the success of recent meta-learning methods, 
we introduced a system that trains multiple models, each 
adapted to a specific class distribution, but requires
only a single set of parameters, as each model is 
derived from a base model by analytic classifier 
adaptation.
The adaptation parameters are extrapolated for the next data distribution based on the data observed so far.
Our experiments on the large-scale \texttt{geo-tweets} 
dataset show that this process results in improved 
prediction quality compared to common baselines, especially when judged by~the~worst-case measure of the lowest accuracy across a day. 

Despite the promising results, the problem of continual 
learning from data streams with time-varying distribution
is far from solved. In particular, our results suggest
that new methods are required to tackle the problem 
that classifiers for certain data distributions are 
harder to learn than others. This suggests that one
might have to adapt not only the model parameters, 
but also the model architecture. 
We plan to address this aspect in future work. 

\bibliographystyle{unsrtnat}
\bibliography{ref}

\begin{thebibliography}{28}
\providecommand{\natexlab}[1]{#1}
\providecommand{\url}[1]{\texttt{#1}}
\expandafter\ifx\csname urlstyle\endcsname\relax
  \providecommand{\doi}[1]{doi: #1}\else
  \providecommand{\doi}{doi: \begingroup \urlstyle{rm}\Url}\fi

\bibitem[Bartlett(1992)]{bartlett1992learning}
Peter~L Bartlett.
\newblock Learning with a slowly changing distribution.
\newblock In \emph{Workshop on Computational Learning Theory (COLT)}, 1992.

\bibitem[Graepel et~al.(2010)Graepel, Candela, Borchert, and
  Herbrich]{graepel2010web}
Thore Graepel, Joaquin~Quinonero Candela, Thomas Borchert, and Ralf Herbrich.
\newblock Web-scale {B}ayesian click-through rate prediction for sponsored
  search advertising in microsoft's bing search engine.
\newblock In \emph{International Conference on Machine Learing (ICML)}, 2010.

\bibitem[McMahan et~al.(2013)McMahan, Holt, Sculley, Young, Ebner, Grady, Nie,
  Phillips, Davydov, Golovin, et~al.]{mcmahan2013ad}
H~Brendan McMahan, Gary Holt, David Sculley, Michael Young, Dietmar Ebner,
  Julian Grady, Lan Nie, Todd Phillips, Eugene Davydov, Daniel Golovin, et~al.
\newblock Ad click prediction: a view from the trenches.
\newblock In \emph{Conference on Knowledge Discovery and Data Mining (KDD)},
  2013.

\bibitem[Li et~al.(2015)Li, Lu, Mei, Wang, and Pandey]{li2015click}
Cheng Li, Yue Lu, Qiaozhu Mei, Dong Wang, and Sandeep Pandey.
\newblock Click-through prediction for advertising in {T}witter timeline.
\newblock In \emph{Conference on Knowledge Discovery and Data Mining (KDD)},
  2015.

\bibitem[Vapnik(2013)]{vapnik2013nature}
Vladimir Vapnik.
\newblock \emph{The nature of statistical learning theory}.
\newblock Springer Science \& Business Media, 2013.

\bibitem[Murshed et~al.(2021)Murshed, Murphy, Hou, Khan, Ananthanarayanan, and
  Hussain]{murshed2021machine}
MG~Sarwar Murshed, Christopher Murphy, Daqing Hou, Nazar Khan, Ganesh
  Ananthanarayanan, and Faraz Hussain.
\newblock Machine learning at the network edge: A survey.
\newblock \emph{ACM Computing Surveys (CSUR)}, 54\penalty0 (8):\penalty0 1--37,
  2021.

\bibitem[Giraud-Carrier(2000)]{giraud2000note}
Christophe Giraud-Carrier.
\newblock A note on the utility of incremental learning.
\newblock \emph{AI Communications}, 13\penalty0 (4):\penalty0 215--223, 2000.

\bibitem[Krawczyk et~al.(2017)Krawczyk, Minku, Gama, Stefanowski, and
  Wo{\'z}niak]{krawczyk2017ensemble}
Bartosz Krawczyk, Leandro~L Minku, Joao Gama, Jerzy Stefanowski, and Micha{\l}
  Wo{\'z}niak.
\newblock Ensemble learning for data stream analysis: A survey.
\newblock \emph{Information Fusion}, 37:\penalty0 132--156, 2017.

\bibitem[Street and Kim(2001)]{street2001streaming}
W~Nick Street and YongSeog Kim.
\newblock A streaming ensemble algorithm {(SEA)} for large-scale
  classification.
\newblock In \emph{Conference on Knowledge Discovery and Data Mining (KDD)},
  pages 377--382, 2001.

\bibitem[Minku and Yao(2011)]{minku2011ddd}
Leandro~L Minku and Xin Yao.
\newblock {DDD:} a new ensemble approach for dealing with concept drift.
\newblock \emph{IEEE Transactions on Knowledge and Data Engineering},
  24\penalty0 (4):\penalty0 619--633, 2011.

\bibitem[Muhlbaier and Polikar(2007)]{muhlbaier2007multiple}
Michael~D Muhlbaier and Robi Polikar.
\newblock Multiple classifiers based incremental learning algorithm for
  learning in nonstationary environments.
\newblock In \emph{International Conference on Machine Learning and
  Cybernetics}, volume~6, 2007.

\bibitem[Cruz et~al.(2018)Cruz, Sabourin, and Cavalcanti]{dsc_survey}
Rafael~M.O. Cruz, Robert Sabourin, and George~D.C. Cavalcanti.
\newblock Dynamic classifier selection: Recent advances and perspectives.
\newblock \emph{Information Fusion}, 41:\penalty0 195--216, 2018.

\bibitem[Almeida et~al.(2018)Almeida, Oliveira, Britto, and
  Sabourin]{dsc_drift}
Paulo~R.L. Almeida, Luiz~S. Oliveira, Alceu~S. Britto, and Robert Sabourin.
\newblock Adapting dynamic classifier selection for concept drift.
\newblock \emph{Expert Systems with Applications}, 104:\penalty0 67--85, 2018.

\bibitem[Rebuffi et~al.(2017)Rebuffi, Kolesnikov, Sperl, and
  Lampert]{rebuffi2017icarl}
Sylvestre-Alvise Rebuffi, Alexander Kolesnikov, Georg Sperl, and Christoph~H
  Lampert.
\newblock {iCarl}: {Incremental} classifier and representation learning.
\newblock In \emph{Conference on Computer Vision and Pattern Recognition
  (CVPR)}, 2017.

\bibitem[Li and Hoiem(2017)]{li2017learning}
Zhizhong Li and Derek Hoiem.
\newblock Learning without forgetting.
\newblock \emph{IEEE Transactions on Pattern Analysis and Machine Intelligence
  (TPAMI)}, 40\penalty0 (12):\penalty0 2935--2947, 2017.

\bibitem[Delange et~al.(2021)Delange, Aljundi, Masana, Parisot, Jia, Leonardis,
  Slabaugh, and Tuytelaars]{delange2021continual}
Matthias Delange, Rahaf Aljundi, Marc Masana, Sarah Parisot, Xu~Jia, Ales
  Leonardis, Greg Slabaugh, and Tinne Tuytelaars.
\newblock A continual learning survey: Defying forgetting in classification
  tasks.
\newblock \emph{IEEE Transactions on Pattern Analysis and Machine Intelligence
  (TPAMI)}, 2021.

\bibitem[McCloskey and Cohen(1989)]{mccloskey1989catastrophic}
Michael McCloskey and Neal~J Cohen.
\newblock Catastrophic interference in connectionist networks: The sequential
  learning problem.
\newblock In \emph{Psychology of learning and motivation}, volume~24, pages
  109--165. Elsevier, 1989.

\bibitem[French(1999)]{french1999catastrophic}
Robert~M French.
\newblock Catastrophic forgetting in connectionist networks.
\newblock \emph{Trends in cognitive sciences}, 3\penalty0 (4):\penalty0
  128--135, 1999.

\bibitem[He et~al.(2020)He, Mao, Shao, and Zhu]{he2020incremental}
Jiangpeng He, Runyu Mao, Zeman Shao, and Fengqing Zhu.
\newblock Incremental learning in online scenario.
\newblock In \emph{Conference on Computer Vision and Pattern Recognition
  (CVPR)}, 2020.

\bibitem[Rusu et~al.(2016)Rusu, Rabinowitz, Desjardins, Soyer, Kirkpatrick,
  Kavukcuoglu, Pascanu, and Hadsell]{rusu2016progressive}
Andrei~A Rusu, Neil~C Rabinowitz, Guillaume Desjardins, Hubert Soyer, James
  Kirkpatrick, Koray Kavukcuoglu, Razvan Pascanu, and Raia Hadsell.
\newblock Progressive neural networks.
\newblock \emph{arXiv preprint arXiv:1606.04671}, 2016.

\bibitem[Yoon et~al.(2018)Yoon, Yang, Lee, and Hwang]{yoon2018lifelong}
Jaehong Yoon, Eunho Yang, Jeongtae Lee, and Sung~Ju Hwang.
\newblock Lifelong learning with dynamically expandable networks.
\newblock In \emph{International Conference on Learning Representations
  (ICLR)}, 2018.

\bibitem[Saerens et~al.(2002)Saerens, Latinne, and
  Decaestecker]{saerens2002adjusting}
Marco Saerens, Patrice Latinne, and Christine Decaestecker.
\newblock Adjusting the outputs of a classifier to new a priori probabilities:
  a simple procedure.
\newblock \emph{Neural Computation}, 14\penalty0 (1):\penalty0 21--41, 2002.

\bibitem[Royer and Lampert(2015)]{royer2015classifier}
Amelie Royer and Christoph~H Lampert.
\newblock Classifier adaptation at prediction time.
\newblock In \emph{Conference on Computer Vision and Pattern Recognition
  (CVPR)}, 2015.

\bibitem[Qui{\~n}onero-Candela et~al.(2008)Qui{\~n}onero-Candela, Sugiyama,
  Schwaighofer, and Lawrence]{quinonero2008dataset}
Joaquin Qui{\~n}onero-Candela, Masashi Sugiyama, Anton Schwaighofer, and Neil~D
  Lawrence.
\newblock \emph{Dataset shift in machine learning}.
\newblock MIT Press, 2008.

\bibitem[Finn et~al.(2017)Finn, Abbeel, and Levine]{maml}
Chelsea Finn, P.~Abbeel, and Sergey Levine.
\newblock {Model-agnostic meta-learning for fast adaptation of deep networks}.
\newblock In \emph{International Conference on Machine Learing (ICML)}, 2017.

\bibitem[Finn et~al.(2019)Finn, Rajeswaran, Kakade, and Levine]{finn2019online}
Chelsea Finn, Aravind Rajeswaran, Sham Kakade, and Sergey Levine.
\newblock Online meta-learning.
\newblock In \emph{International Conference on Machine Learing (ICML)}, 2019.

\bibitem[Reimers and Gurevych(2019)]{reimers-2019-sentence-bert}
Nils Reimers and Iryna Gurevych.
\newblock {Sentence-BERT}: Sentence embeddings using siamese {BERT}-networks.
\newblock In \emph{Conference on Empirical Methods on Natural Language
  Processing (EMNLP)}, 2019.

\bibitem[Han et~al.(2014)Han, Cook, and Baldwin]{han2014text}
Bo~Han, Paul Cook, and Timothy Baldwin.
\newblock Text-based {T}witter user geolocation prediction.
\newblock \emph{Journal of Artificial Intelligence Research}, 49:\penalty0
  451--500, 2014.

\end{thebibliography}

\end{document}